\documentclass[10pt,twocolumn,letterpaper]{article}

\usepackage[pagenumbers]{cvpr} 

\definecolor{cvprblue}{rgb}{0.21,0.49,0.74}
\usepackage[pagebackref,breaklinks,colorlinks,allcolors=cvprblue]{hyperref}
\usepackage[ruled,vlined]{algorithm2e}
\usepackage{algorithmic}
\usepackage{multirow}
\usepackage{arydshln}
\def\paperID{14193} 
\def\confName{CVPR}
\def\confYear{2025}

\title{ZipVL: Efficient Large Vision-Language Models with Dynamic Token Sparsification}

\author{%
  Yefei He\textsuperscript{1,2} \thanks{Work done during an internship at Shanghai AI Laboratory.}
  \quad Feng Chen\textsuperscript{3}
  \quad Jing Liu\textsuperscript{4}
  \quad Wenqi Shao\textsuperscript{2} \\[0.1cm]\hspace{-1.25em}
  \quad  Hong Zhou\textsuperscript{1}$^{\dagger}$
  \quad  Kaipeng Zhang\textsuperscript{2}\thanks{Corresponding authors.}
  \quad  Bohan Zhuang\textsuperscript{1}
  \\[0.2cm]
  \textsuperscript{1}Zhejiang University, China \\
  \textsuperscript{2}Shanghai AI Laboratory, China \\
  \textsuperscript{3}The University of Adelaide, Australia \\
  \textsuperscript{4}ZIP Lab, Monash University, Australia
}

\begin{document}
\maketitle
\begin{abstract}
The efficiency of large vision-language models (LVLMs) is constrained by the computational bottleneck of the attention mechanism during the prefill phase and the memory bottleneck of fetching the key-value (KV) cache in the decoding phase, particularly in scenarios involving high-resolution images or videos. 
Visual content often exhibits substantial redundancy, resulting in highly sparse attention maps within LVLMs. This sparsity can be leveraged to accelerate attention computation or compress the KV cache through various approaches.
However, most studies focus on addressing only one of these bottlenecks and do not adequately support dynamic adjustment of sparsity concerning distinct layers or tasks.
In this paper, we present ZipVL, an efficient inference framework designed for LVLMs through a dynamic ratio allocation strategy of important tokens. 
This ratio is adaptively determined based on the layer-specific distribution of attention scores, rather than fixed hyper-parameters, thereby improving efficiency for less complex tasks while maintaining high performance for more challenging ones.
Then we select important tokens based on their normalized attention scores and perform sparse attention mechanism solely on those important tokens, reducing the latency in the prefill phase. 
Tokens deemed less important will be discarded to reduce KV cache size, alleviating the memory bottleneck in the decoding phase.
Our experiments demonstrate that ZipVL can accelerate the prefill phase by 2.3$\times$ and improve decoding throughput by 2.8$\times$, with a minimal accuracy reduction of only 0.5\% on VQAv2 benchmark over LLaVA-Next-13B model, effectively enhancing the generation efficiency of LVLMs.
\end{abstract}    
\section{Introduction}
\label{sec:intro}

With the recent advancement of large language models (LLMs)~\cite{achiam2023gpt4, team2023gemini,vavekanand2024llama3}, many studies have extended their capabilities to comprehend and generate visual content. These models, commonly known as large vision-language models (LVLMs), have demonstrated remarkable performance in tasks such as image captioning and visual question answering~\cite{gemaking,liu2024visual,team2024chameleon,ge2024seedx,lin2023videollava}. Typically, to remain compatible with the next-token-prediction generation scheme of LLMs, images or videos are encoded into visual tokens through a pre-trained visual encoder, and concatenated with text tokens for input into the model. 
However, for high-resolution images or videos, the visual encoder generates excessive sequences of visual tokens, significantly limiting the generative efficiency of LVLMs. 
For example, a short video consisting of 128 frames is encoded into over 18,000 tokens by the LongVA model~\cite{zhang2024longva}.
In such cases, the prefill phase suffers from the quadratic complexity of the attention mechanism, resulting in \textbf{computational bottleneck} and prolonged time-to-first-token (TTFT). In the decoding phase, each new token interacts with all preceding tokens, requiring to fetch the full key-value (KV) cache from memory and thereby inducing a \textbf{memory bottleneck}. Improving generative efficiency in both phases is essential for the practical deployment of LVLMs.

\begin{figure*}[t]
    \centering
    \includegraphics[width=0.95\linewidth]{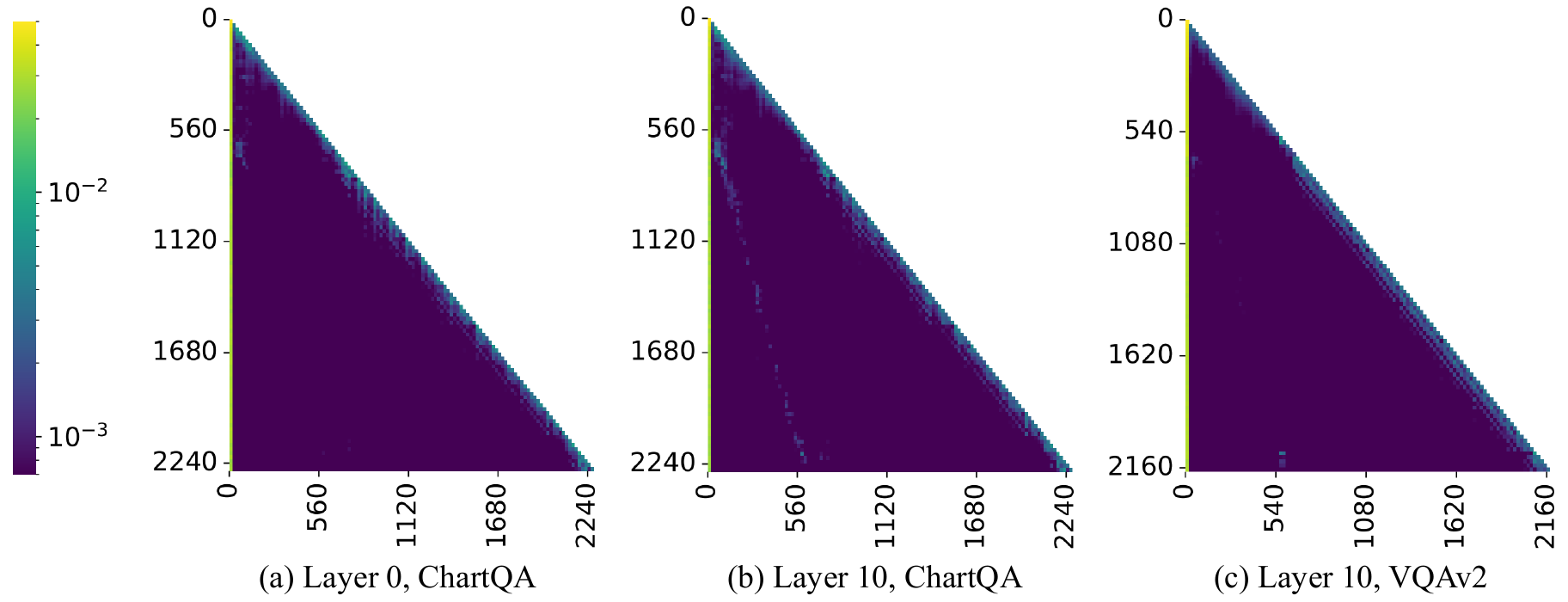}
    \caption{
    The attention maps exhibit distinct sparse patterns across different layers (subfigures (a) and (b)) and vary significantly between tasks (subfigures (b) and (c)). Data was collected from the LLaVA-Next-7B model using input samples from the VQAv2 and ChartQA datasets.
    }
    \vspace{-1em}
    \label{fig:attention_variation}
\end{figure*}

To address computational complexity in the prefill phase, sparse attention~\cite{pagliardini2023dynamicsparse,jiangminference,zhu2024sampleattention} has emerged as an effective strategy, particularly suitable for LVLMs where visual information exhibits considerable redundancy, resulting in higher sparsity in attention maps compared to LLMs~\cite{wan2024lookm, chen2024fastv}. This sparsity can be implemented at various levels of granularity. Some studies pre-define several sparse patterns and assign them to the attention mask during inference~\cite{jiangminference,zhu2024sampleattention}. However, these predefined patterns are incompatible with efficient attention implementations such as FlashAttention~\cite{dao2022flashattention} and require custom GPU kernels for each pattern. Alternatively, other approaches adopt token-level sparsity by identifying and discarding less important tokens~\cite{chen2024fastv, arif2024hired}, allowing seamless integration with off-the-shelf efficient attention implementations. However, the optimal retention ratio of important tokens may vary across different layers or tasks due to distinct attention patterns, as illustrated in Figure~\ref{fig:attention_variation}. These methods rely on a fixed token retention ratio and do not dynamically adjust based on task difficulty, leading to suboptimal performance on complex tasks.

To alleviate memory bottleneck in the decoding phase, various efforts have been made to reduce KV cache size, including token dropping~\cite{wan2024lookm}, token merging~\cite{yang2024pyramidinfer}, and quantization~\cite{hooper2024kvquant,he2024zipcache}. However, these methods often rely on fixed compression ratios that are uniformly applied across all layers, failing to account for the distinct characteristics of attention maps in different layers. Moreover, despite the necessity of identifying important tokens for both sparse attention and KV cache compression, a unified inference optimization framework has yet to be developed.

\begin{figure*}[t]
    \centering
    \includegraphics[width=0.95\linewidth]{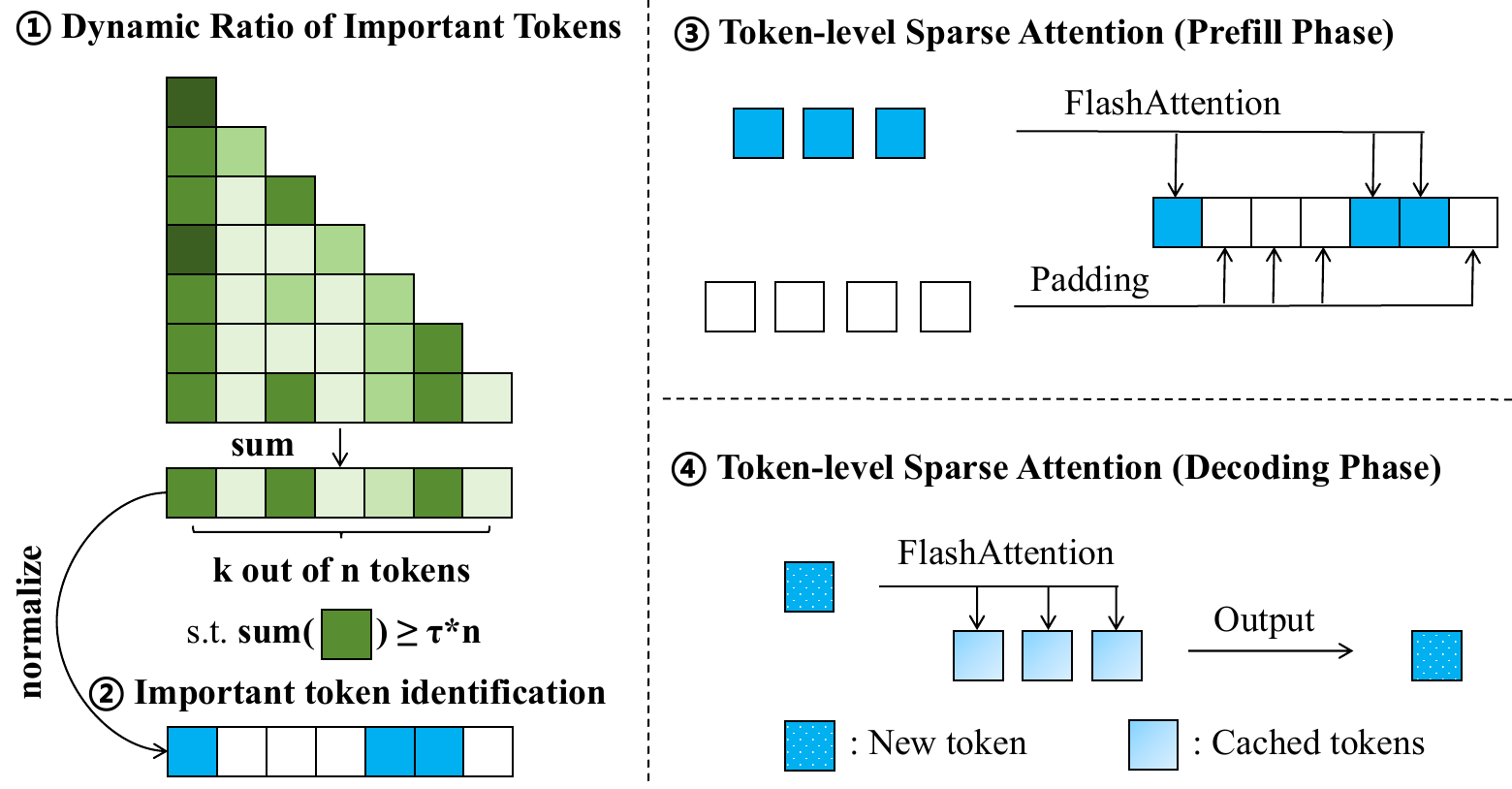}
    \caption{
    Overview of the proposed ZipVL framework during the prefill phase. Here, $\tau$ represents the threshold for retaining attention scores, $n$ and $p$ are the total number of tokens and the number of important tokens, respectively. After determining the ratio of important tokens and identifying them, we optimize both prefill and decoding phases by exclusively computing attention for important tokens. The KV cache for tokens deemed less important will be discarded.
    }
    \label{fig:overview}
\end{figure*}

In this paper, we present ZipVL, an efficient inference framework tailored for LVLMs that jointly optimizes the prefill and decoding phases with a unified ratio of important tokens, as shown in Figure~\ref{fig:overview}. To start with, we introduce a layer-wise adaptive ratio assignment scheme for important tokens. This ratio is adaptively determined based on the distribution of attention scores in each layer, rather than relying on predefined hyper-parameters~\cite{chen2024fastv,arif2024hired,he2024zipcache,zhang2023h2o}. This adaptive approach allows the ratio to be adjusted according to task complexity, enhancing efficiency for simpler tasks while preserving performance for more complex ones. After determining the ratio, we then select important tokens with the highest normalized attention scores, following prior work~\cite{he2024zipcache, ren2024efficacy}. To alleviate the \textbf{computational bottleneck} in the prefill phase, sparse attention is performed at the token level by computing attention only for the selected important tokens. Notably, this approach seamlessly integrates with existing fast attention implementations without requiring custom GPU kernels. To tackle the \textbf{memory bottleneck}, the same set of important tokens is applied to compress the KV cache, evicting tokens deemed less important. Extensive experiments on multimodal benchmarks demonstrate that our method achieves nearly lossless performance while reducing prefill phase latency by $2.3\times$ and improving decoding throughput by $2.8\times$.

In summary, our contributions are as follows:

\begin{itemize}[leftmargin=*]
\item We propose an adaptive layer-wise ratio assignment scheme for important tokens. The ratio is dynamically determined based on the distribution of attention scores and varies across different layers and tasks, enhancing performance and efficiency compared to a fixed ratio scheme.

\item We introduce a unified approach to jointly optimize the prefill and decoding stages through the adaptive allocation of important tokens. Tokens considered less important are excluded from attention computation to reduce computational complexity, and their KV cache will be discarded to alleviate the memory bottleneck in the decoding phase.

\item By integrating these techniques, we present ZipVL, an efficient inference framework tailored for LVLMs. Comprehensive experiments across diverse benchmarks validate the efficacy of ZipVL, demonstrating that it achieves state-of-the-art performance in both accuracy and generation efficiency for LVLMs.
\end{itemize}

\section{Related Work}
\label{sec:related}
\subsection{Sparse Attention}
Attention scores have been widely observed to exhibit high sparsity in both LLMs and LVLMs~\cite{xiaoefficientsink,wan2024lookm,zhu2024sampleattention,zaheer2020bigbird,beltagy2020longformer}. This sparsity allows sparse attention to overcome the quadratic computational complexity of the standard attention mechanism by restricting each token to focus on only a subset of tokens within the input sequence~\cite{zhu2024sampleattention,jiangminference,pagliardini2023dynamicsparse,ribarsparq}. 
Depending on the granularity of sparsity, sparse attention can be categorized into unstructured, semi-structured, and structured schemes. 
The unstructured scheme~\cite{xFormers2022,pytorch2024flexattention} employs sparse attention masks without a fixed structure, making it hardware-unfriendly and challenging to achieve practical inference acceleration. The semi-structured sparse attention uses attention masks with predefined sparse patterns~\cite{jiangminference, pagliardini2023dynamicsparse,zhu2024sampleattention} or introduces N:M sparsity to attention weights~\cite{chen2023dynamic}. However, it requires customized computational kernels for each sparse pattern or specific hardware to achieve acceleration. Structured sparse attention \cite{chen2024fastv,arif2024hired} directly prunes tokens before the attention computation, enabling acceleration without the need for custom kernels. However, due to its coarse granularity, the pruning sparsity and the selection of tokens to prune significantly impact model performance. For instance, HiRED~\cite{arif2024hired} selects patches with the highest responses based on the feature maps of the visual encoder without considering the input text prompt, leading to suboptimal performance. FastV~\cite{chen2024fastv} empirically retains all tokens in the first two layers and prunes 50\% of the visual tokens in all subsequent layers, resulting in performance degradation in challenging tasks such as ChartQA~\cite{masry2022chartqa}. In contrast, our approach achieves superior performance through an adaptive layer-wise ratio assignment scheme for important tokens.

\subsection{KV Cache Compression}
KV cache prevents re-computation in the decoding phase by storing the key and value states of previous tokens, but with a significant memory bottleneck in long-context scenarios. Previous efforts to compress the KV cache can be broadly categorized into three types: token dropping-based~\cite{gemodel,ren2024efficacy,zhang2023h2o}, token merging-based~\cite{wang2024modeltells, wan2024lookm, liu2024efficient}, and quantization-based approaches~\cite{hooper2024kvquant,he2024zipcache,yang2024notoken,kang2024gear,liu2024kivi}. In this paper, we reduce the KV cache size by evicting tokens deemed less important. The proposed layer-wise adaptive ratio assignment scheme boosts the overall compression ratio and accuracy compared to a fixed ratio for important tokens. Additionally, it should be noted that our method can be combined with quantization-based approaches to achieve further reductions in KV cache size.
\section{Preliminary}
Attention block is the key module of Transformer-based LLMs. Each attention block contains three weight matrices $\mathbf{W}_{\mathrm{Q}}, \mathbf{W}_{\mathrm{K}}, \mathbf{W}_{\mathrm{V}} \in \mathbb{R}^{d\times d}$, where $d$ is the dimension of the input data. Here, we use a single attention head and omit the output projection for clarity. In the prefill phase, the input data $\mathbf{X} \in \mathbb{R}^{n\times d}$ with a sequence length of $n$ is first multiplied with three weight matrices to obtain the query, key and value states:
\begin{equation} \label{eq:qkv_prefill}
    \mathbf{Q} = \mathbf{X} \mathbf{W}_{\mathrm{Q}}, \quad \mathbf{K} = \mathbf{X} \mathbf{W}_{\mathrm{K}}, \quad \mathbf{V} = \mathbf{X} \mathbf{W}_{\mathrm{V}}.    
\end{equation}

Then the attention output is calculated as follows:
\begin{gather} \label{eq:attention_score}
\mathbf{A} = \mathrm{Softmax}\left(\frac{\mathbf{Q}\mathbf{K}^T+\mathbf{M}}{\sqrt{d}}\right), \mathbf{O}=\mathbf{A }\mathbf{V}.
\end{gather}
Here, computing the product of $\mathbf{Q}\mathbf{K}^T$ has a quadratic complexity $O ({n}^ 2)$, which makes the prefill phase \textbf{compute-bound}. $\mathbf{M} \in \mathbb{R}^{n\times n}$ is a lower triangular causal mask to ensure that each token can only attend to itself and previous tokens. Unstructured and semi-structured sparse attention introduce sparsity in the attention mask $\mathbf{M}$ with dynamic or fixed sparse pattern. With custom computing kernels, tokens in certain positions can be skipped when computing $\mathbf{Q}\mathbf{K}^T$, thus accelerating the computation. On the other hand, structured sparse attention only computes attention scores for a subset of tokens $\mathbf{X}'\in \mathbb{R}^{n' \times d} $, reducing computational complexity to $O ({n'}^ 2)$ and seamlessly integrating with existing fast attention implementations.

For the decoding phase, the input data is the embedding of the current token $\mathbf{x} \in \mathbb{R}^{1 \times d}$. To enable the interaction between the current token and all previous tokens, the KV cache of previous tokens needs to be fetched from memory, making the decoding phase \textbf{memory-bound}:
\begin{equation} \label{eq:qkv_decoding}
    \mathbf{q} = \mathbf{x} \mathbf{W}_{\mathrm{Q}}, \quad
    \mathbf{k} = \mathbf{x} \mathbf{W}_{\mathrm{K}}, \quad
    \mathbf{v} = \mathbf{x} \mathbf{W}_{\mathrm{V}}.
    \end{equation}
    \begin{equation} \label{eq:retrive_kvcache}
    \mathbf{K} = \mathrm{Concat}(\mathbf{K}, \mathbf{k}), \quad \mathbf{V} = \mathrm{Concat}(\mathbf{V}, \mathbf{v}).
\end{equation}
The attention outputs are then computed as follows with a computational complexity of $O (n)$:
\begin{gather} \label{eq:attention_decode}
\mathbf{a} = \mathrm{Softmax}\left(\frac{\mathbf{q}\mathbf{K}^T}{\sqrt{d}}\right), \quad \mathbf{o}=\mathbf{a} \mathbf{V}.
\end{gather}
\section{Method}
\subsection{Layer-wise Adaptive Ratio Assignment for Important tokens}
Prior studies~\cite{arif2024hired, zhang2023h2o, he2024zipcache,liu2024kivi,wan2024lookm} typically adopt a fixed ratio of important tokens across all layers. However, as analyzed by the preceding study~\cite{chen2024fastv} and demonstrated in Figure~\ref{fig:attention_variation}(a) and (b), there are substantial variations in the attention map patterns across different layers. Moreover, Figure~\ref{fig:attention_variation}(b) and (c) illustrate that, even within the same layer, attention maps can differ depending on the task and input. In scenarios involving complex tasks, a limited, static ratio for important tokens can impair model performance. This raises the question: 
\begin{center}
    \textit{can the model dynamically determine the number of tokens required to solve a task?}
\end{center}

Intuitively, for simpler tasks, the model needs to concentrate on fewer tokens, leading to a more focused distribution of attention scores. Conversely, more demanding tasks require the model to engage with a broader array of tokens, resulting in a more uniform distribution of attention scores. Prior work~\cite{xiaoefficientsink} also highlights the criticality of preserving significant attention scores during inference within a constrained attention window. Building on these insights, we introduce a layer-wise adaptive scheme for assigning ratio of important tokens, ensuring the majority of significant attention scores are maintained within each layer.

Consider an attention layer with $n$ input tokens, where the full attention score matrix is denoted as $\mathbf{A} \in \mathbb{R}^{n \times n}$. The accumulated attention score for each token $j$ is calculated by summing the corresponding column: 
\begin{gather} \label{eq:accumulated_attention}
a_j=\sum_{c=1}^{n} \mathbf{A}_{c,j}.
\end{gather}

\begin{figure}
    \centering
    \includegraphics[width=0.95\linewidth]{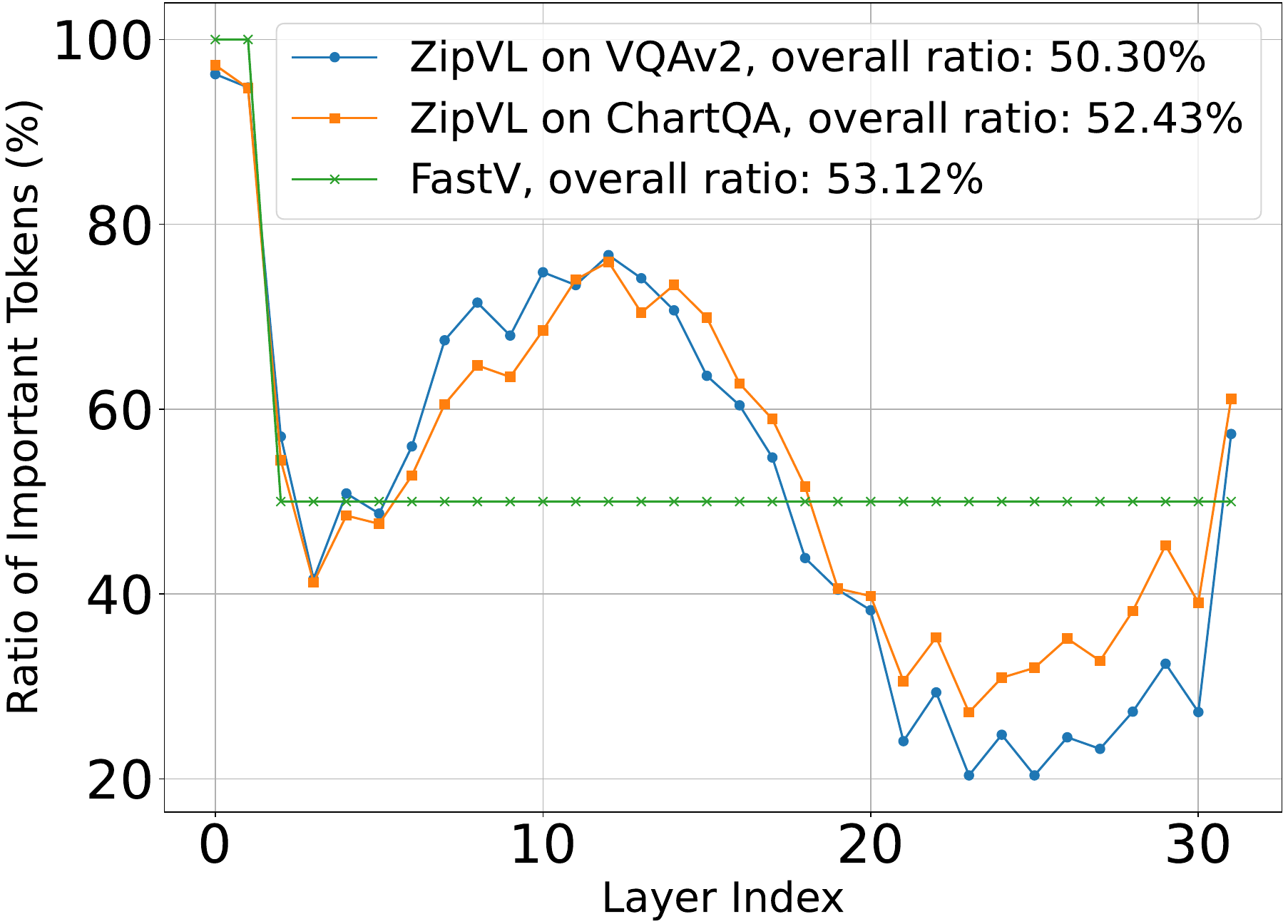}
    \caption{The ratio of important tokens distributed across layers. Data was collected from the LLaVA-Next-7B model using input samples from the VQAv2 and ChartQA datasets with a $\tau$ of 0.975.}
    \label{fig:layerwise_budget}
\end{figure}

\begin{figure*}[htbp]  
\centering  
\begin{subfigure}{.31\linewidth}   \caption{LLaVA-v1.5-7B}
  \centering  
  \includegraphics[width=\linewidth]{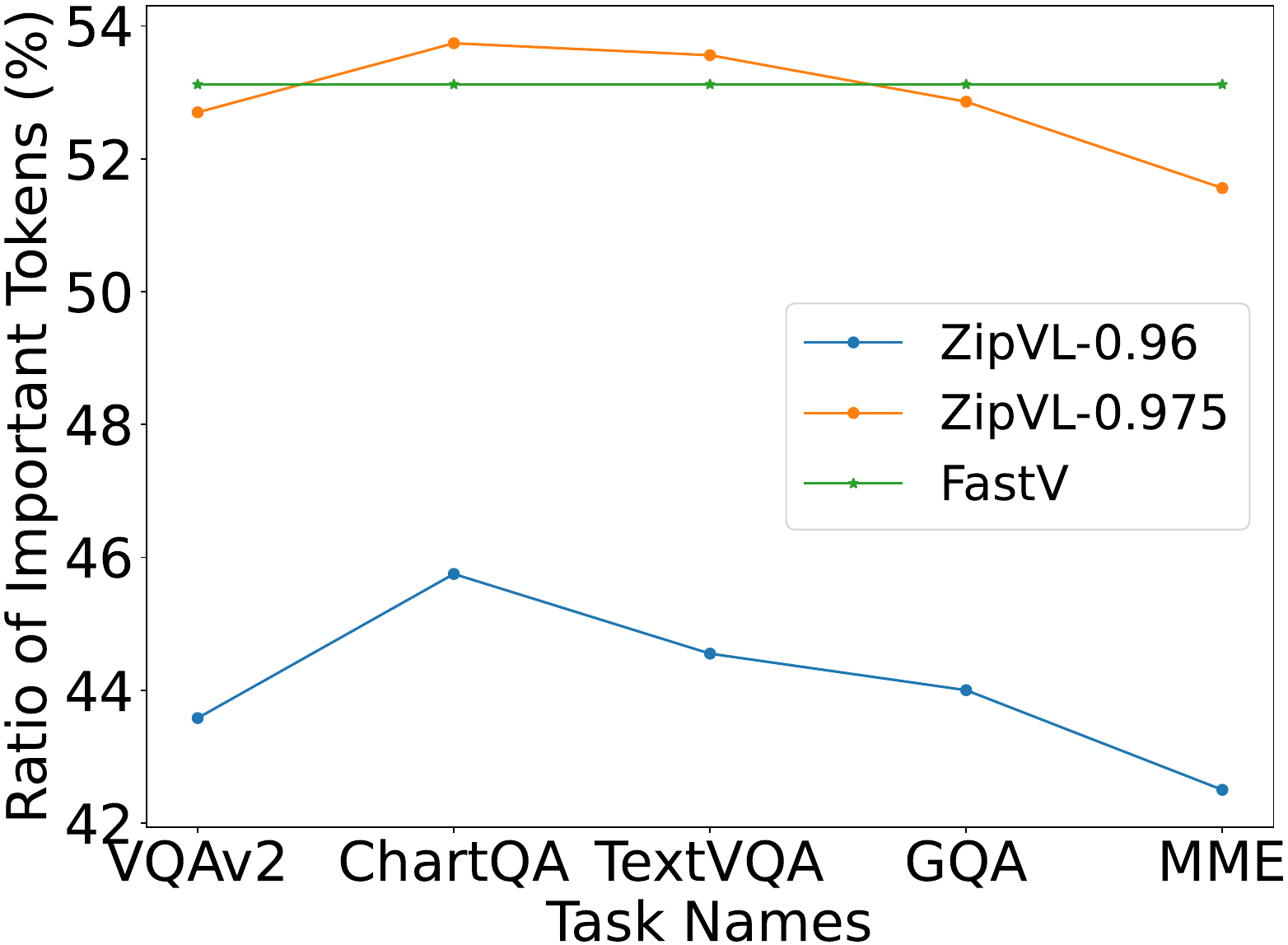}  
  \label{fig:sub1}  
\end{subfigure}%
  \begin{subfigure}{.321\linewidth}  \caption{LLaVA-Next-7B}
  \centering  
  \includegraphics[width=\linewidth]{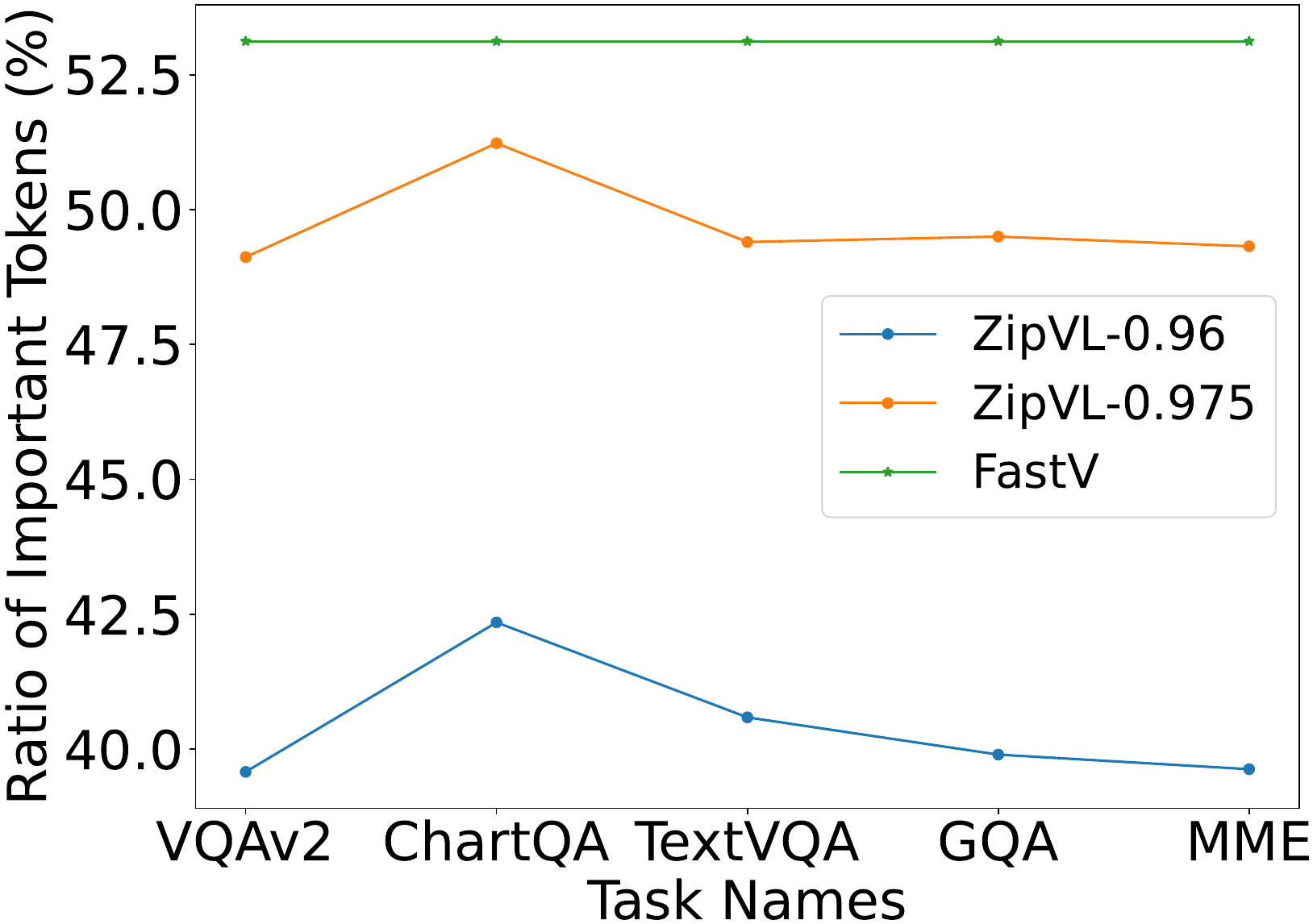}  
  \label{fig:sub2}  
\end{subfigure}%
\begin{subfigure}{.31\linewidth}  \caption{LLaVA-Next-13B}
  \centering  
  \includegraphics[width=\linewidth]{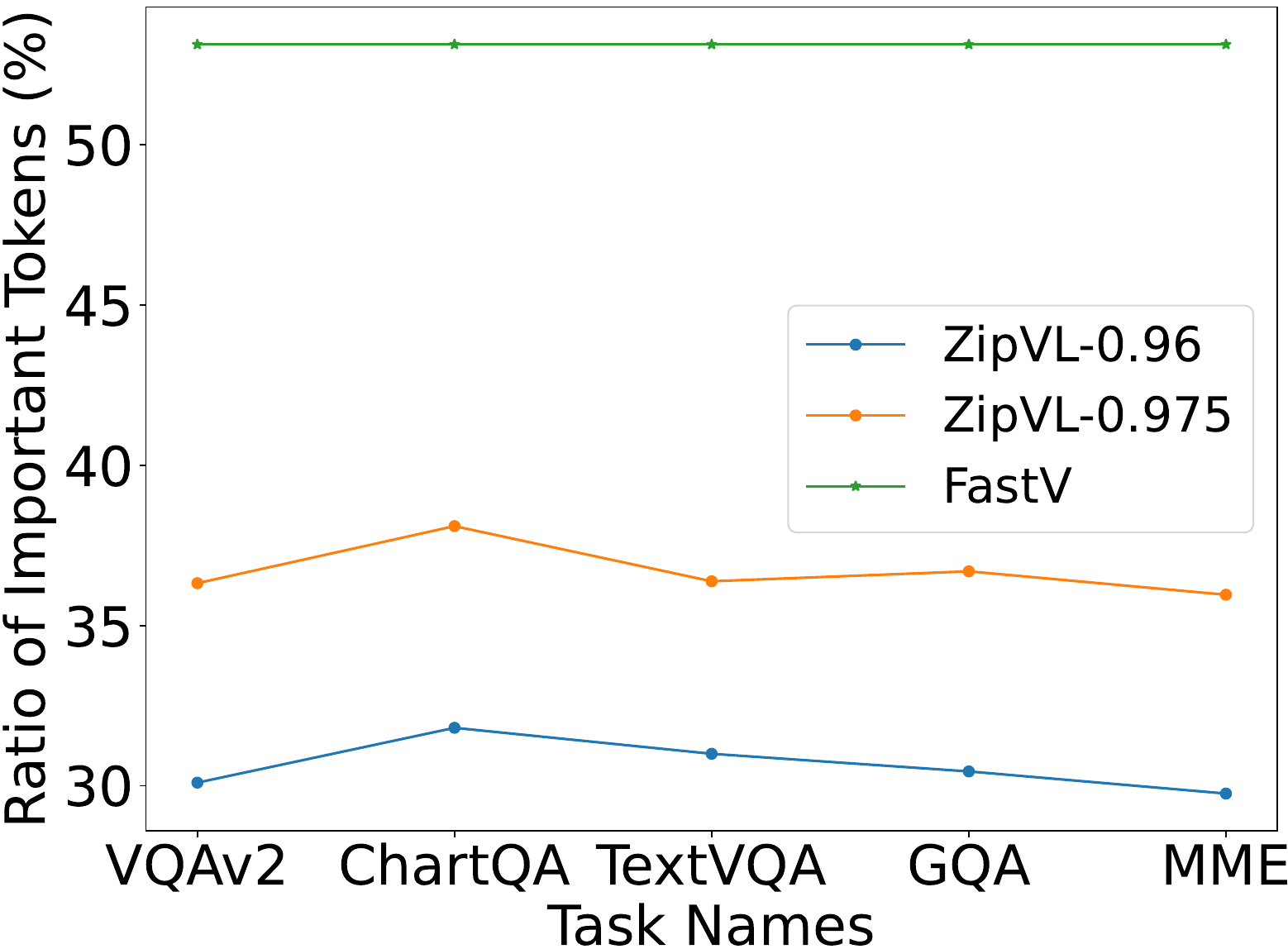}  
  \label{fig:sub3}  
\end{subfigure}%
\caption{The ratio of important tokens across different methods on different tasks. The proposed ZipVL can adaptively determine this ratio based on the attention scores, assigning more ratio to important tokens on complex tasks.}  
\label{fig:budget_size}  
\end{figure*}

These accumulated attention scores are subsequently sorted in descending order, such that $a_{\text{sorted}(j)}$ represents the $j$-th highest attention score.  The number of important tokens $p$ is determined by preserving the majority of attention scores with a minimal number of tokens, which can be expressed as:
\begin{equation} \label{eq:determine_budget}
    p=\mathrm{min}\{p \in \mathbb{Z} \mid \sum_{j=1}^{p} a_{\text{sorted}(j)} \geq \tau \times n\}.
\end{equation}
Here, $\tau$ is the threshold dictating the retention of attention scores and the sum of the attention scores in $\mathbf{A}$ is equal to $n$ due to the row-wise $\mathrm{Softmax}$ operation. As shown in Figure~\ref{fig:layerwise_budget}, our method can dynamically adjust the ratio of important tokens across distinct layers and tasks, thereby enhancing performance in complex tasks while improving efficiency in simpler tasks. Additional experimental results can be found in Section~\ref{sec:image_results} and Figure~\ref{fig:budget_size}.

\subsection{Inference Optimization with Unified Important Token Allocation} \label{sec:optimization}
 After determining the number of important tokens $p$ for each layer, we partition all tokens into two sets: set $\mathbf{T}$ of important tokens with a size of $p$ and the set $\mathbf{U}$ for less important tokens with a size of $n-p$. Following prior work~\cite{ren2024efficacy, he2024zipcache}, we use normalized attention scores to assess token importance, calculated as follows:
\begin{gather} \label{eq:mean_attention}
\tilde{a}_j = \frac{\sum_{c=1}^{n} \mathbf{A}_{c,j}}{\text{nnz}(\mathbf{A}_{:,j})}.
\end{gather}
Here, $\text{nnz}(\mathbf{A}_{:,j})$ denotes the number of non-zero elements in the $j$-th column. Important tokens are then selected using the top-$k$ indexing method, while the remainder are considered less important:
\begin{gather} \label{eq:important_token}
\mathbf{T}=\mathrm{topk\_index}( \tilde{a}_j, p),
\end{gather}
\begin{equation} \label{eq:unimportant_token}
    \mathbf{U} = \{ j \in \{1, 2, \dots, n\} \mid j \notin \mathbf{T} \}.
\end{equation}
The inference optimization is then performed based on the split of tokens. Specifically, to address the \textbf{computational bottleneck} in the prefill phase, the attention mechanism is performed solely on these important tokens, thereby enhancing efficiency through token-level sparsity. Tokens excluded from this computation have their outputs padded to maintain the number of tokens consistent for subsequent layers. 
By leveraging token-level sparsity, our approach seamlessly integrates with off-the-shelf, fast attention implementations~\cite{dao2022flashattention} to expedite the prefill process.

To mitigate the \textbf{memory bottleneck}, we retain the KV cache for important tokens while discarding the less important ones. During the decoding phase, the current token interacts solely with the cached important tokens, thereby reducing the memory overhead of fetching the KV cache for computation.

\noindent \textbf{Efficient approximation of full attention scores.} Full attention scores are not accessible in fast attention implementation~\cite{dao2022flashattention}. In such case, to circumvent the computation of full attention scores in Eqs.~(\ref{eq:accumulated_attention}) and~(\ref{eq:mean_attention}), we selectively compute and accumulate the attention scores for a subset of tokens, following previous literature~\cite{he2024zipcache, jiangminference}. The obtained partial attention scores are then used as the approximation of the full attention scores. The size of this subset is small and fixed, ensuring that the computational burden for these tokens remains minimal in long-context scenarios. The accumulated and normalized attention scores for each token can then be approximated with partial attention scores. Details and ablation experiments on this approximation can be found in the supplementary material.

Overall, the attention mechanisms for the prefill and decoding phases are summarized in Algorithms~\ref{algo:zipvl_prefill} and \ref{algo:zipvl_decoding}, respectively.

\begin{algorithm}[t]
  \SetKwProg{myProcedure}{procedure}{\string:}{end procedure}
  \SetKwProg{myFunc}{function}{\string:}{end function}
  \myProcedure{\FuncSty{ZipVL Prefill}}{
    \KwIn{Input embedding $\mathbf{X}$, threshold $\tau$}
    \KwOut{Attention output $\mathbf{O}$, KV cache ($\mathbf{K}, \mathbf{V}$) }

    Calculate query, key and value states ($\mathbf{Q}$,$\mathbf{K}$,$\mathbf{V}$) as per Eq.~(\ref{eq:qkv_prefill})
    
    Select a subset of tokens $\mathbf{Q}'$ from query states and compute attention scores $\mathbf{A}'=\mathrm{Softmax}\left(\mathbf{Q}'\mathbf{K}^T\right)$
    
    Determine the number of important tokens as per Eq.~(\ref{eq:determine_budget})
    
    Calculate the normalized attention scores for each token as per Eq.~(\ref{eq:mean_attention})
    
    Select a set of important tokens $\mathbf{T}$ as per Eq.~(\ref{eq:important_token})
    
    \tcp{Token-level Sparse Attention with FlashAttention}
    $\mathbf{O}=\text{FlashAttention}(\mathbf{Q[\mathbf{T}]},\mathbf{K[\mathbf{T}]},\mathbf{V[\mathbf{T}]})$ \\
    
    \tcp{KV Cache}
    $\mathbf{K}=\mathbf{K[\mathbf{T}]}$\\
    $\mathbf{V}=\mathbf{V[\mathbf{T}]}$\\
    
    \Return{$\mathbf{O}$, ($\mathbf{K}$, $\mathbf{V}$)}
  }
    \caption{The prefill phase of ZipVL}
  \label{algo:zipvl_prefill}
\end{algorithm}

\begin{algorithm}[t]
  \SetKwProg{myProcedure}{procedure}{\string:}{end procedure}
  \SetKwProg{myFunc}{function}{\string:}{end function}
  \myProcedure{\FuncSty{ZipVL Decoding}}{
      \KwIn{Input embedding $\mathbf{x}$, stored KV cache ($\mathbf{K}_{\mathrm{in}}, \mathbf{V}_{\mathrm{in}}$)}
      \KwOut{Attention output $\mathbf{o}$, updated KV cache ($\mathbf{K}_{\mathrm{out}}, \mathbf{V}_{\mathrm{out}}$)}

      Calculate query, key and value states ($\mathbf{q}$,$\mathbf{k}$,$\mathbf{v}$) as per Eq.~(\ref{eq:qkv_prefill})
      
      Fetch KV cache from memory: $\mathbf{K}_{\mathrm{out}} = \mathrm{Concat}(\mathbf{K}_{\mathrm{in}}, \mathbf{k}), \quad \mathbf{V}_{\mathrm{out}} = \mathrm{Concat}(\mathbf{V}_{\mathrm{in}}, \mathbf{v})$
      
      Compute attention output $\mathbf{o}=\text{FlashAttention}(\mathbf{q},\mathbf{K}_{\mathrm{out}},\mathbf{V}_{\mathrm{out}})$

      \Return{$\mathbf{o}$, ($\mathbf{K}_{\mathrm{out}}$, $\mathbf{V}_{\mathrm{out}}$)}
      
  } 
  \caption{The decoding phase of ZipVL}
  \label{algo:zipvl_decoding}
\end{algorithm}
\begin{table*}[h] 
\caption{Performance comparisons of image LVLMs on various benchmarks. Here, ``Ratio'' denotes the proportion of tokens participating in attention computation. ``$\dagger$'' denotes token-level sparsity is only employed in attention modules.}
\label{tab:main_image}
\centering
\scalebox{0.96}{
\begin{tabular}{cccccccc}
\hline
Model                           & Method              & Ratio & VQAv2         & ChartQA       & TextVQA       & GQA           & MME           \\ \hline
\multirow{6}{*}{LLaVA-v1.5-7B}  & Full                & 100\%  & 76.6          & 18.2          & 46.1          & 61.9          & 1507          \\
                                & FastV$\dagger$               & 53.1\% & 75.8          & 17.7          & 45.5          & 60.2          & \textbf{1511}          \\
                                & HiRED               & 20\%   & 73.0          & 17.3          & 45.6          & 56.8          & 1368          \\
                                & HiRED               & 40\%   & 75.5          & 17.6          & 45.6          & 59.5          & 1433          \\
                                & Ours ($\tau$=0.96)  & 44.1\% & 76.1          & \textbf{18.0}          & 45.0          & 61.3          & 1495          \\
                                & Ours ($\tau$=0.975) & 52.8\% & \textbf{76.5} & 17.9 & \textbf{45.7} & \textbf{61.7} & 1505 \\ \hline
\multirow{6}{*}{LLaVA-Next-7B}  & Full                & 100\%  & 80.3          & 54.8          & 64.8          & 64.1          & 1519          \\
                                & FastV$\dagger$               & 53.1\% & 79.5          & 51.2          & 63.7          & 63.7          & 1490          \\
                                & HiRED               & 20\%   & 77.5          & 42.0          & 61.4          & 61.4          & 1483          \\
                                & HiRED               & 40\%   & 78.8          & 46.5          & 61.8          & 59.4          & 1474          \\
                                & Ours ($\tau$=0.96)  & 40.4\% & 79.4          & 51.0          & 62.8          & 63.8          & 1473          \\
                                & Ours ($\tau$=0.975) & 49.3\% & \textbf{79.8} & \textbf{52.6} & \textbf{64.0} & \textbf{64.0} & \textbf{1497} \\ \hline
\multirow{6}{*}{LLaVA-Next-13B} & Full                & 100\%  & 80.9          & 66.2          & 66.9          & 65.7          & 1570          \\
                                & FastV$\dagger$               & 53.1\% & 76.8          & 51.6          & 59.7          & 62.9          & 1555          \\
                                & HiRED               & 20\%   & 77.9          & 48.9          & 63.6          & 63.1          & 1545          \\
                                & HiRED               & 40\%   & 79.3          & 53.7          & \textbf{65.2} & 64.1          & 1570 \\
                                & Ours ($\tau$=0.96)  & 30.6\% & 79.7          & 57.8          & 63.6          & 64.3          & 1551          \\
                                & Ours ($\tau$=0.975) & 36.8\% & \textbf{80.4} & \textbf{58.5} & 64.9          & \textbf{64.8} & \textbf{1574}   \\ \hline
\end{tabular} }
\end{table*}

\section{Experiments}
\subsection{Implementation Details}
To assess the effectiveness of our proposed method, we conduct experiments on both image and video understanding tasks. For image understanding, we utilize three widely adopted LVLMs: LLaVA~\cite{lin2023videollava}, LLaVA-Next~\cite{liu2024llavanext}, and QWen-VL~\cite{Qwen-VL}. These models are evaluated against five rigorous benchmarks: VQAv2~\cite{goyal2017vqav2}, TextVQA~\cite{singh2019textvqa}, GQA~\cite{hudson2019gqa}, MME~\cite{fu2023mme}, and ChartQA~\cite{masry2022chartqa}. For video understanding, evaluations are conducted using the LongVA~\cite{zhang2024longva} model on the Video-MME~\cite{fu2024videomme} benchmark.
To ensure reproducibility, all reported results are obtained using the Evaluation Suite of Large Multimodal Models~\cite{lmms_eval2024}. 

\subsection{Main Results}
\subsubsection{Evaluation on Image Benchmarks} \label{sec:image_results}
We begin our evaluation on five image comprehension benchmarks and compare our results against well-established methods with token-level sparsity: FastV~\cite{chen2024fastv} and HiRED~\cite{arif2024hired}. The results are presented in Table~\ref{tab:main_image}. Notably, HiRED determines the importance of patches through the feature map of the visual encoder, without considering the semantic information of the input prompt, resulting in a significant accuracy drop. In contrast, both FastV and our approach assess token importance via attention maps in the LVLMs. However, FastV employs a fixed token ratio and exhibits severe performance degradation on challenging tasks such as ChartQA~\cite{masry2022chartqa}. By implementing layer-wise adaptive ratio assignment, our proposed ZipVL consistently surpasses FastV across all five benchmarks and three model architectures, while maintaining a smaller overall ratio of important tokens. As illustrated in Figure~\ref{fig:budget_size}, our method dynamically adjusts the ratio across various tasks and models, slightly increasing the ratio of important tokens for difficult tasks to preserve performance while enhancing efficiency on simpler tasks. Moreover, the performance gap between our method and FastV becomes more pronounced over the LLaVA-Next-13B model. This discrepancy can be attributed to the varying attention maps across different models and that FastV's predefined hyperparameters are not universally applicable, whereas our dynamic approach demonstrates high robustness. 

\begin{table*}[h] 
\caption{Performance comparisons of video LVLMs on Video-MME benchmark. Here, ``Attn FLOPs Reduction'' denotes the reduction in floating-point operations (FLOPs) of the attention mechanism. ``$\dagger$'' denotes token-level sparsity is only employed in attention modules.}
\label{tab:main_video}
\centering
\scalebox{0.96}{
\begin{tabular}{ccccccccc}
\hline
Model                       & Frames               & Method             & \begin{tabular}[c]{@{}c@{}}Attn FLOPs\\ Reduction (\%)\end{tabular} & \begin{tabular}[c]{@{}c@{}}KV Cache\\ Reduction (\%)\end{tabular} & Short         & Medium        & Long          & Overall       \\ \hline
\multirow{10}{*}{LongVA-7B} & \multirow{5}{*}{64}  & Full               & 0              & 0                                               & 61.4          & 50.9          & 45.0          & 52.4          \\
                            &                      & QK-sparse          & 47.0        & 0                                                  & 60.9          & 51.4          & \textbf{45.1} & 52.4          \\
                            &                      & MInference         & 54.2           & 0                                               & 60.7          & 51.2          & 44.6          & 52.1          \\
                            &                      & FastV$\dagger$     & 71.7         & 46.4                                                 & 61.0          & 50.6          & 45.0          & 52.2          \\
                            &                      & Ours($\tau$=0.975) & \textbf{77.0}      & \textbf{52.0}                                                    & \textbf{61.0} & \textbf{51.4} & 45.0          & \textbf{52.4} \\ \cline{2-9} 
                            & \multirow{5}{*}{128} & Full               & 0            & 0                                                & 61.1          & 50.4          & 46.2          & 52.6          \\
                            &                      & QK-sparse          & 46.9      & 0                                                  & \textbf{61.3} & 49.7          & \textbf{46.3} & 52.4          \\
                            &                      & MInference         & 77.1       & 0                                                 & 61.0          & 50.5          & 45.3          & 52.3          \\
                            &                      & FastV$\dagger$     & 71.7        & 46.4                                                 & 60.2          & 50.2          & 46.2          & 52.2          \\
                            &                      & Ours($\tau$=0.975) & \textbf{82.3}     & \textbf{57.9}                                                    & 61.1          & \textbf{50.5} & 46.1          & \textbf{52.6} \\ \hline
\end{tabular} }
\end{table*}
\subsubsection{Evaluation on Video Benchmarks}
We also assess the performance of our method on the Video-MME benchmark \cite{fu2024videomme} over the LongVA model \cite{zhang2024longva}, which supports a maximum multimodal input length of 224K tokens. We compare our approach with semi-structured sparse attention methods such as MInference \cite{jiangminference} and QK-sparse \cite{pagliardini2023dynamicsparse}, as well as the structured sparse attention method FastV \cite{chen2024fastv}. The results are summarized in Table~\ref{tab:main_video}. Among these sparse attention methods, FastV \cite{chen2024fastv} consistently retains a fixed proportion of tokens while MInference \cite{jiangminference} retains a fixed number of sparse blocks. Notably, our approach not only achieves the highest overall performance but also exhibits superior reductions in FLOPs within the attention module and KV cache size compared to other sparse attention methods. This demonstrates the effectiveness of employing dynamic token-level sparsity to accelerate the attention module in LVLMs. Furthermore, long videos inherently contain significant redundancy, and our method dynamically allocates the ratio of important tokens by analyzing the sparse attention maps, resulting in a higher FLOPs reduction ratio when processing 128-frame videos compared to 64-frame videos.

\subsection{Ablation Study} \label{sec:ablation}
\subsubsection{Effect of the Layer-wise Adaptive Ratio} \label{sec:ablation_layeradaptive}
In this subsection, we evaluate the efficacy of the proposed adaptive ratio assignment scheme by integrating it with sparse attention and KV cache compression, as detailed in Table~\ref{tab:ablation}. Initially, we implement a fixed sparse attention scheme on LongVA-7B model over Video-MME benchmark. In this scheme, the ratio for important tokens remains constant across all attention layers and is fixed. Although this approach shares the same overall important token ratio and FLOPs reduction ratio as our method, it suffers from significant performance degradation (51.1\% vs. 52.6\%) due to its failure to account for the varying attention maps across layers. In contrast, our method achieves nearly lossless performance (52.4\% vs. 52.6\%) while reducing the FLOPs of attention mechanism by 82.3\%.

To ablate the efficacy of our method for KV cache compression, we integrate the adaptive ratio assignment scheme with the mixed-precision KV cache quantization approach~\cite{he2024zipcache} and evaluate its performance over LLaMA3-8B~\cite{llama3} model on the GSM8k dataset. For both methods, important tokens are quantized to 4-bit, while other tokens are quantized to 2-bit. Notably, by adaptively determining the ratio of important tokens for each layer, our method achieves a significantly higher compression ratio (6.18 $\times$ vs. 4.69$\times$) while maintaining superior accuracy (54.06\% vs. 53.75\%). This demonstrates that our method also sets a new state-of-the-art for \textbf{KV cache compression of LLMs}.

\begin{table}[h] 
\caption{The effect of the proposed adaptive ratio assignment scheme on sparse attention and KV cache compression. Here, ``Ratio'' denotes the proportion of important tokens. For Video-MME benchmark, the input videos consist of 128 frames.}
\label{tab:ablation}
\centering
\scalebox{0.75}{
\begin{tabular}{cccc}
\hline
\multicolumn{4}{c}{\textbf{Sparse Attention}}            \\ \hline
Method & Ratio (\%) & Attn FLOPs Reduction (\%)   & Video-MME (\%) \\
LongVA-7B   & 100      & 0                 & 52.6         \\
Fixed  & 42.1      & 82.3                 & 51.1         \\
Ours   & 42.1      & 82.3                 & \textbf{52.4}         \\ \hline
\multicolumn{4}{c}{\textbf{KV Cache Compression}}        \\ \hline
Method & Ratio (\%) & Compression Ratio & GSM8k Acc. (\%) \\
LLaMA3-8B   & 100      & 1$\times$                 & 55.88         \\
Fixed~\cite{he2024zipcache}  & 70.0      & 4.69$\times$                 & 53.75         \\
Ours   & \textbf{28.6}      & \textbf{6.18$\times$}                 & \textbf{54.06}         \\ \hline
\end{tabular} }
\end{table}

\subsubsection{Effect of the Threshold $\tau$}
\begin{figure}[t]
    \centering
    \includegraphics[width=0.88\columnwidth]{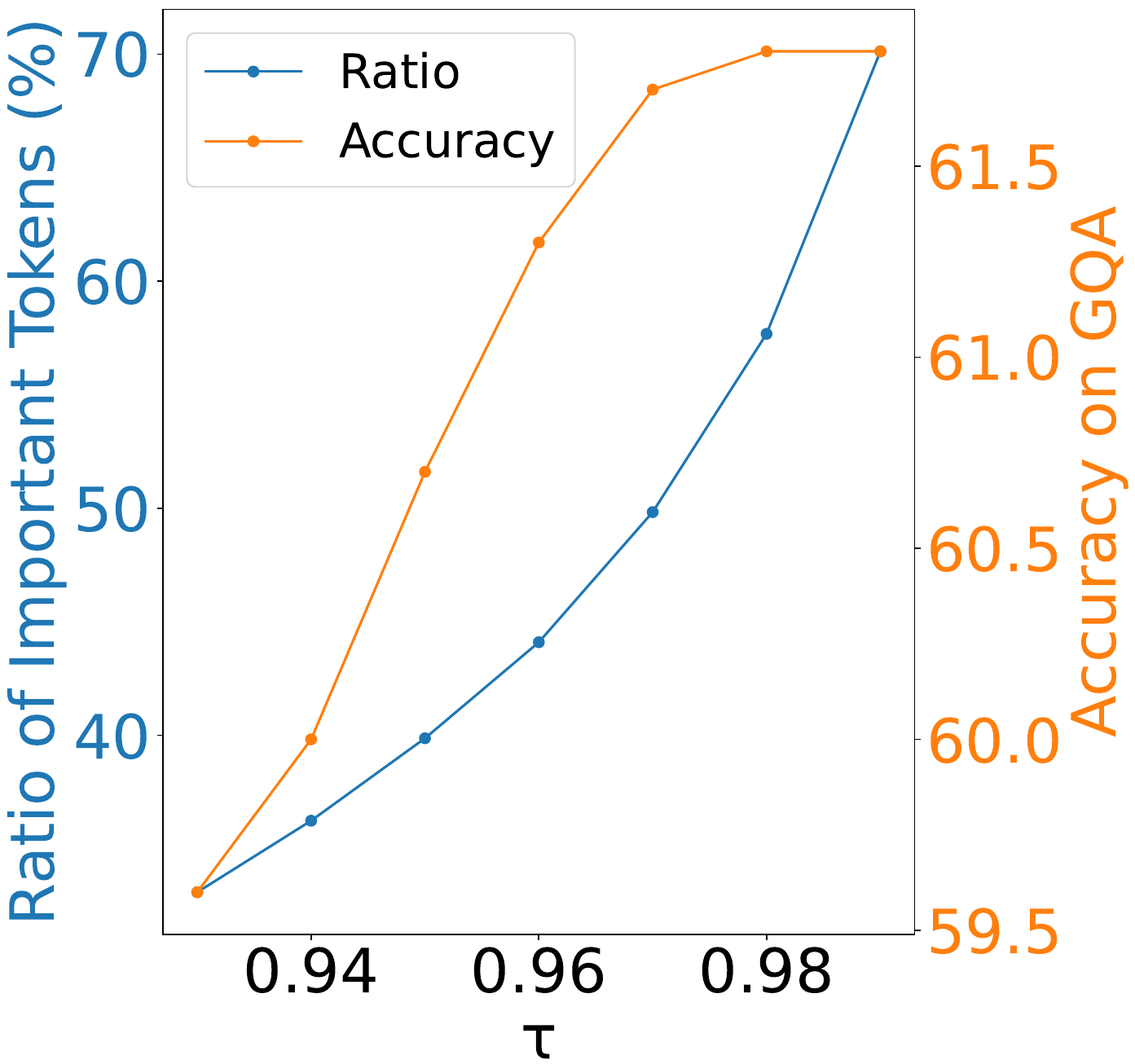}
    \caption{
    The effect of attention scores retention threshold $\tau$ on the ratio of important tokens and the model performance. Data was collected on GQA benchmark over LLaVA-v1.5-7B model.
    }
    \vspace{-1em}
    \label{fig:tau_ablation}
\end{figure}

We further investigate the impact of the attention retention threshold $\tau$ on both the ratio of important tokens and model performance. The results are illustrated in Figure~\ref{fig:tau_ablation}. Intuitively, a lower retention threshold leads to a reduced ratio for important tokens, thereby enhancing generation efficiency at the cost of performance degradation. Notably, the ratio decreases significantly as $\tau$ decreases but remains above 0.97, with minimal performance deterioration. Conversely, when $\tau$ falls below 0.97, substantial performance loss is observed, despite a gradual reduction in the ratio of important tokens. This indicates that the optimal range for $\tau$ lies around 0.97.

Ablation experiments on the effect of $\tau$ across other models and benchmarks, as well as on probe token selection and token importance metrics, are provided in the supplementary material.

\subsection{Deployment Efficiency}
In this subsection, we present the prefill phase latency and decoding throughput in Table~\ref{tab:efficiency} to illustrate the real efficiency improvements achieved by ZipVL. Specifically, we first compare the prefill phase latency of ZipVL with that of the well-established semi-structured sparse attention method, MInference~\cite{jiangminference}. Notably, MInference exhibits significant additional overhead when the sequence length is short and is notably slower than FlashAttention~\cite{dao2022flashattention} for sequence lengths below 32K. In contrast, ZipVL achieves comparable latency to FlashAttention with short input sequences, while significantly reducing the prefill phase latency as the sequence length exceeds 32K. This can be attributed to the fact that the attention module's latency becomes the dominant factor in the total latency with long sequences. With an input sequence length of 128K, ZipVL achieves a 2.3$\times$ reduction in prefill-phase latency.

Moreover, MInference is not designed to reduce the KV cache size, while the proposed ZipVL jointly optimizes the attention computation and the KV cache through the dynamic token ratio. 
With the reduction in KV cache size, ZipVL can perform inference with a larger batch size.
Consequently, ZipVL presents a 2.8$\times$ higher decoding throughput compared to baseline method~\cite{dao2022flashattention} with an input sequence length of 16K.

Additional results of latency breakdown can be found in the supplementary material.

\begin{table}[]
\caption{Comparisons of prefill phase latency and decoding throughput across different sequence lengths. Data is collected from LLaVA-Next-13B model on an Nvidia A100 GPU. Here, ``TTFT'' denotes time-to-first-token and is measured with a batch size of 1. ``OOM'' indicates out-of-memory during the decoding process. Throughput is measured using the maximum batch size that can be supported by a single GPU.}
\label{tab:efficiency}
\centering
\scalebox{0.95}{
\begin{tabular}{cccc}
\hline
Input Length          & Method         & \begin{tabular}[c]{@{}c@{}}TTFT$\downarrow$\\ (s)\end{tabular} & \begin{tabular}[c]{@{}c@{}}Throughput$\uparrow$\\ (token/s)\end{tabular} \\ \hline
\multirow{3}{*}{8K}   & FlashAttention & 1.33                                                           & 27.79                                                                    \\
                      & MInference     & 3.30                                                           & 27.79                                                                    \\
                      & Ours           & \textbf{1.31}                                                  & \textbf{75.54}                                                           \\ \hline
\multirow{3}{*}{16K}  & FlashAttention & 2.64                                                           & 14.10                                                                    \\
                      & MInference     & 4.57                                                           & 14.10                                                                    \\
                      & Ours           & \textbf{2.61}                                                  & \textbf{40.87}                                                           \\ \hline
\multirow{3}{*}{32K}  & FlashAttention & 5.90                                                           & OOM                                                                      \\
                      & MInference     & 6.51                                                           & OOM                                                                      \\
                      & Ours           & \textbf{5.04}                                                  & \textbf{15.31}                                                           \\ \hline
\multirow{3}{*}{64K}  & FlashAttention & 15.61                                                          & OOM                                                                      \\
                      & MInference     & 10.92                                                          & OOM                                                                      \\
                      & Ours           & \textbf{10.51}                                                 & \textbf{10.29}                                                           \\ \hline
\multirow{3}{*}{128K} & FlashAttention & 48.01                                                          & OOM                                                                      \\
                      & MInference     & \textbf{20.08}                                                 & OOM                                                                      \\
                      & Ours           & 20.57                                                          & OOM                                                                      \\ \hline
\end{tabular}
}
\end{table}
\section{Conclusion and Future Work}
In this paper, we have proposed ZipVL, an efficient inference framework tailored for LVLMs. ZipVL jointly optimizes both the prefill and decoding phases by assigning an adaptive ratio of important tokens. This ratio is dynamically adjusted based on the distribution of attention scores across each layer, ensuring that the majority of attention scores are preserved. After identifying important tokens through normalized attention scores, less significant tokens are excluded from attention computation to alleviate the computational bottleneck. Additionally, their KV cache is evicted, mitigating the memory bottleneck in the decoding phase. Extensive experiments have demonstrated that ZipVL significantly enhances the generation efficiency of LVLMs, achieving up to a 2.3$\times$ reduction in prefill phase latency and a 2.8$\times$ higher decoding throughput.
However, a limitation of our approach is its focus on sparse attention only, while the multi-layer perceptron (MLP) modules in both phases remain dense. Future efforts may explore extending sparse computations to MLP modules to further reduce computational complexity.

\setcounter{section}{0}
\setcounter{equation}{0}
\setcounter{figure}{0}
\setcounter{table}{0}
\renewcommand\thesection{\Alph{section}}
\renewcommand\thefigure{\Alph{figure}}
\renewcommand\thetable{\Alph{table}}
\renewcommand{\theequation}{\Alph{equation}}

\maketitlesupplementary

\section{Efficient Approximation of Full Attention Scores} \label{sec:appendix_approximation}
ZipVL requires accumulated attention scores to adaptively assign the ratio of important tokens and normalized attention scores to identify token importance. However, attention scores are not accessible in fast attention implementations such as FlashAttention~\citep{dao2022flashattention}. To integrate our method with FlashAttention, we follow prior literature~\citep{he2024zipcache, jiangminference} and select a subset of tokens, referred to as ``\textbf{probe tokens}"~\citep{he2024zipcache}, and explicitly compute their attention scores:
\begin{equation} \label{eq:probe_attention}
    \mathbf{A}_{probe} = \mathrm{Softmax}\left(\frac{\mathbf{Q}_{probe}\mathbf{K}^T}{\sqrt{d_k}}\right).
\end{equation}
The approximate accumulated and normalized attention scores for each token can then be obtained accordingly based on $\mathbf{A}_{probe}$. Prior work~\citep{he2024zipcache} selects 10\% of the tokens as probe tokens, which still yields quadratic complexity in Eq.~\ref{eq:probe_attention}. In contrast, we select only 64 recent tokens and 64 randomly positioned tokens, which incurs negligible computation overhead in long-context scenarios.

\section{Additional Experimental Results}
\subsection{Ablation on Probe Tokens}
In this subsection, we conduct ablation studies on the selection of probe tokens, as shown in Table~\ref{tab:ablation_probe}. When 128 randomly positioned tokens are used as probe tokens, the accuracy drops significantly to 6.3\%. While relying solely on recent tokens delivers reasonable performance, a hybrid approach that combines recent and randomly positioned tokens demonstrates superior performance (52.6\% compared to 52.4\%). Notably, this hybrid strategy achieves accuracy comparable to computing full attention scores without approximation, while substantially reducing computational overhead.

\begin{table*}[b]
\caption{Performance comparisons across different probe token selection approaches on ChartQA benchmark. Here, ``Ratio'' denotes the proportion of tokens involved in attention computation. ``All'' denotes all tokens are used as probe tokens, requiring full attention score computation. To ensure a fair comparison, the threshold $\tau$ is adjusted to maintain a similar ``Ratio" across approaches.}
\label{tab:ablation_probe}
\centering
\scalebox{0.99}{
\begin{tabular}{cccccc}
\hline
Model                          & Method                 & Probe Tokens           & $\tau$   & Ratio (\%)    & Acc. (\%)     \\ \hline
\multirow{6}{*}{LLaVA-Next-7B} & Original               & -                      & -     & 100           & 54.8          \\ \cline{2-6} 
                               & \multirow{5}{*}{ZipVL} & 64 recent              & 0.980 & 50.5          & 52.3          \\
                               &                        & 128 recent             & 0.987 & 50.5          & 52.4          \\
                               &                        & 128 random             & 0.975 & 51.3          & 6.3           \\
                               &                        & 64 recent \& 64 random & 0.975 & \textbf{50.6} & \textbf{52.6} \\
                               &                        & All                    & 0.960 & 53.6          & 52.6          \\ \hline
\end{tabular}
}
\end{table*}

\subsection{Ablation on Importance Metric}
This subsection evaluates the impacts of different metrics on layer-wise adaptive ratio assignment and the identification of important tokens, as depicted in Table~\ref{tab:ablation_metric}. For adaptive ratio assignment, accumulated attention scores consistently outperform normalized attention scores in terms of overall performance. Conversely, for the identification of important tokens, employing normalized attention scores yields higher accuracy, which is consistent with the findings of prior studies~\cite{he2024zipcache, ren2024efficacy} in LLMs.

\begin{table*}[]
\caption{Performance comparisons across different metric for adaptive ratio assignment and important token identification. Data is collected over LLaVA-Next-7B model on ChartQA benchmark. Here, ``Ratio'' denotes the proportion of tokens involved in attention computation. To ensure a fair comparison, the threshold $\tau$ is adjusted to maintain similar ``Ratio" values across approaches.}
\label{tab:ablation_metric}
\centering
\scalebox{0.99}{
\begin{tabular}{ccccc}
\hline
Metric for Adaptive Ratio Assignment & Metric for Important Token Identification & $\tau$   & Ratio (\%) & Acc. (\%)      \\ \hline
Accumulated Attention Scores         & Accumulated Attention Scores              & 0.975 & 50.28      & 52.40          \\
Accumulated Attention Scores         & Normalized Attention Scores               & 0.975 & 50.55      & \textbf{52.64} \\
Normalized Attention Scores          & Accumulated Attention Scores              & 0.995 & 51.84      & 51.56          \\
Normalized Attention Scores          & Normalized Attention Scores               & 0.995 & 51.45      & 51.84          \\ \hline
\end{tabular}
}
\end{table*}

\subsection{Effect of the Threshold $\tau$}
\begin{figure*}[b]  
\centering  
\begin{subfigure}{.343\linewidth}   \caption{LLaVA-Next-7B on GQA}
  \centering  
  \includegraphics[width=\linewidth]{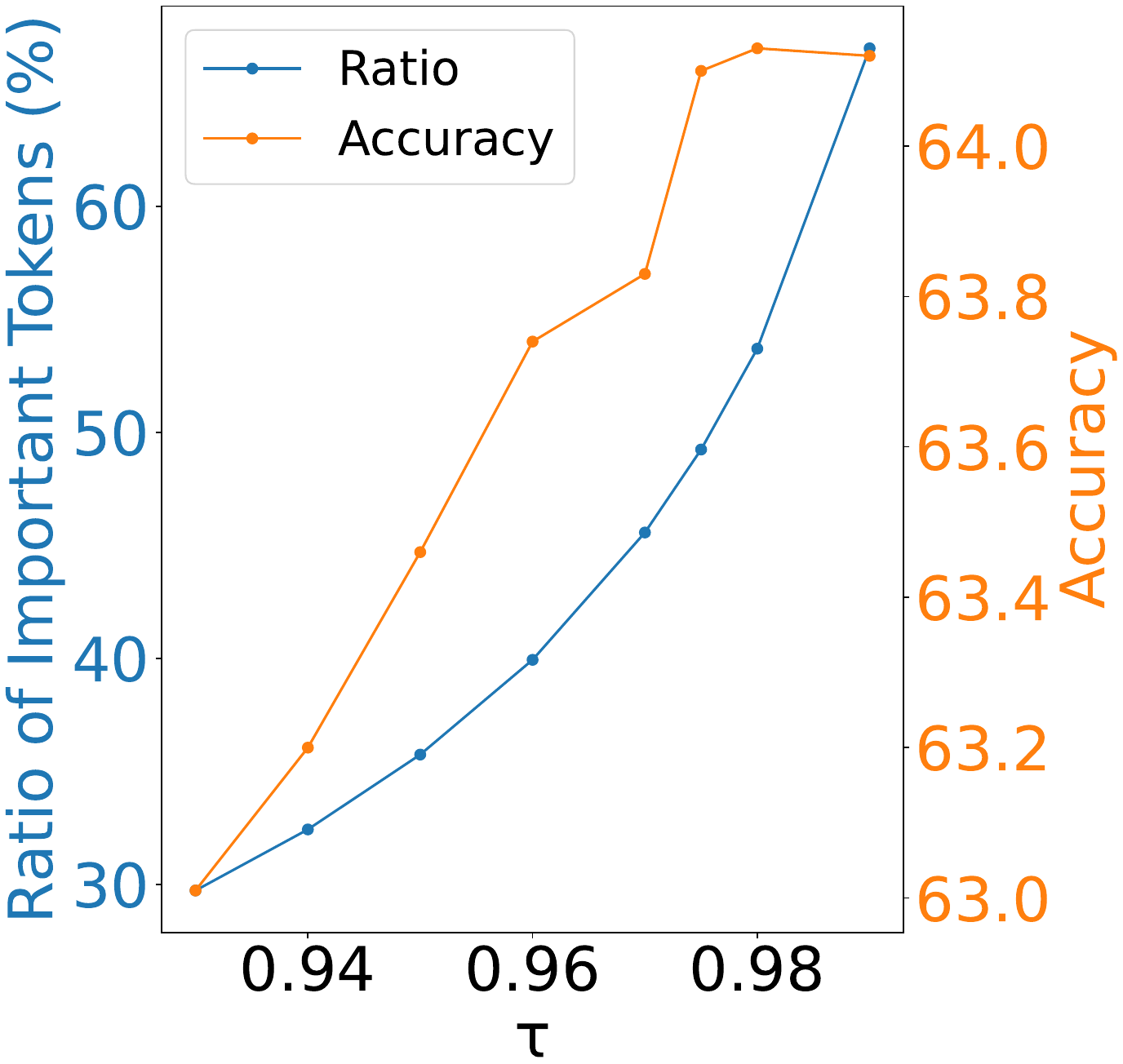}  
  \label{fig:tau_sub1}  
\end{subfigure}%
  \begin{subfigure}{.328\linewidth}  \caption{LLaVA-Next-7B on ChartQA}
  \centering  
  \includegraphics[width=\linewidth]{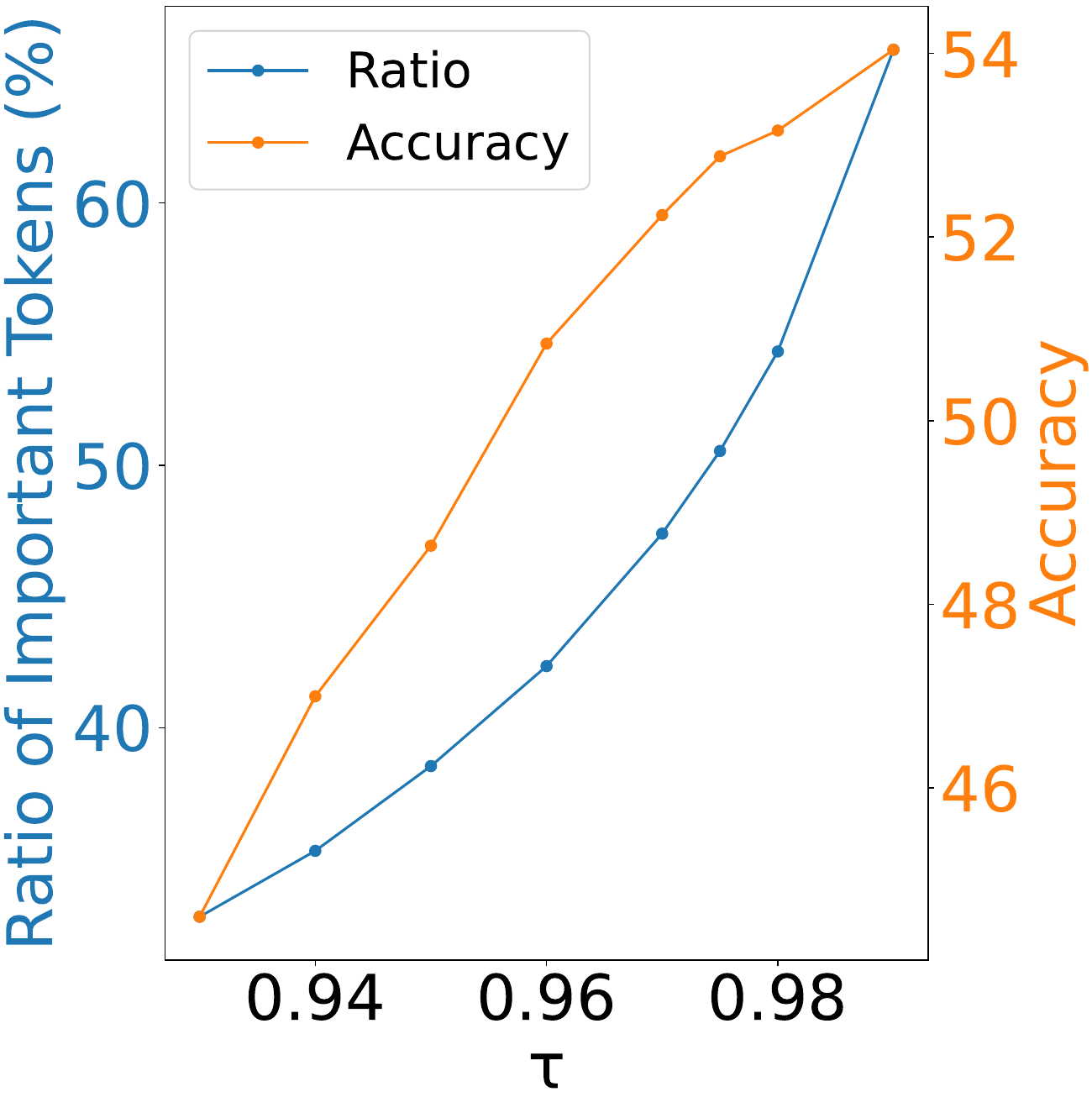}  
  \label{fig:tau_sub2}  
\end{subfigure}%
\begin{subfigure}{.33\linewidth}  \caption{LLaVA-Next-13B on ChartQA}
  \centering  
  \includegraphics[width=\linewidth]{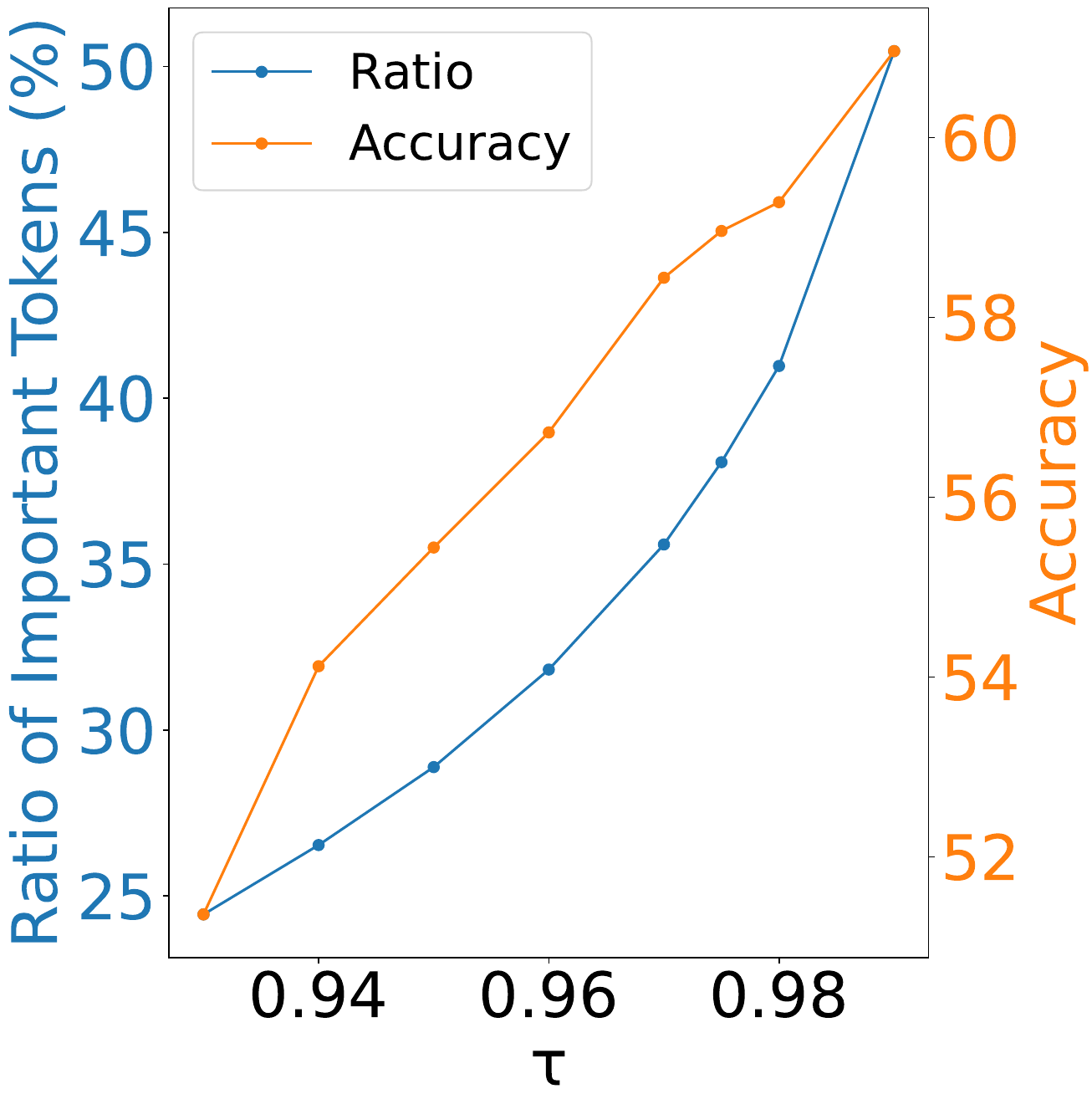}  
  \label{fig:tau_sub3}  
\end{subfigure}%
\caption{The effect of attention scores retention threshold $\tau$ on the ratio of important tokens and the model performance.}  
\label{fig:ablate_tau}  
\end{figure*}

This subsection investigates the impact of the attention retention threshold $\tau$ on the proportion of important tokens and model performance across different models and benchmarks, as shown in Figure~\ref{fig:ablate_tau}. When $\tau$ decreases but stays above 0.98, the proportion of important tokens drops significantly, while the accuracy declines only marginally. This suggests improved generation efficiency with minimal performance loss. However, when $\tau$ falls below 0.97, a noticeable drop in model performance occurs, accompanied by a continued decrease in the proportion of important tokens.

\subsection{Overhead Analysis}
ZipVL introduces additional overhead to adaptively assign the ratio of important tokens and identify token importance. In this subsection, we provide a detailed latency breakdown analysis, as summarized in Table~\ref{tab:overhead}. The primary source of overhead arises from computing approximate attention scores, while other operations, such as sorting and top-$k$ selection, contribute minimal overhead. Nonetheless, the speed gains achieved through sparse attention compensate for these costs, leading to a reduction in the overall time-to-first-token (TTFT). Furthermore, as the input sequence length increases, the impact of overhead diminishes due to the use of a fixed number of probe tokens for approximating attention scores.

\begin{table*}[]
\caption{The overhead of each operation in ZipVL. Here, ``TTFT'' denotes time-to-first-token and is measured with a batch size of 1.}
\label{tab:overhead}
\centering
\scalebox{0.99}{
\begin{tabular}{cccc}
\hline
Model                           & Input Length         & Method                 & TTFT (s) \\ \hline
\multirow{10}{*}{LLaVA-Next-7B} & \multirow{5}{*}{16K} & Original               & 3.48     \\
                                &                      & + approximate attention & 4.38     \\
                                &                      & + sort \& cumsum        & 4.41     \\
                                &                      & + normalize \& top-k    & 4.43     \\
                                &                      & + sparse attention      & \textbf{3.05}     \\ \cline{2-4} 
                                & \multirow{5}{*}{32K} & Original               & 9.35     \\
                                &                      & + approximate attention & 10.95    \\
                                &                      & + sort \& cumsum        & 10.96    \\
                                &                      & + normalize \& top-k    & 11.02    \\
                                &                      & + sparse attention      & \textbf{6.40}     \\ \hline
\end{tabular} }
\end{table*}


{
    \small
    \bibliographystyle{ieeenat_fullname}
    \bibliography{main}
}


\end{document}